\documentclass[runningheads]{llncs}
\usepackage{algorithmic}
\usepackage{graphicx}
\usepackage{textcomp}
\usepackage{xcolor}
\usepackage{caption}
\usepackage{amssymb}
\usepackage{dsfont}
\usepackage{multirow}
\usepackage{makecell}
\usepackage{multirow}
\usepackage{subcaption}

\begin{document}
	\title{Automated discovery of trade-off between utility, privacy and fairness in machine learning models}
	\titlerunning{Discovery of trade-off between utility, privacy and fairness in ML}
	%
	\author{Bogdan Ficiu\inst{1}
		\and
		Neil D. Lawrence\inst{2}\orcidID{0000-0001-9258-1030}
		\and
		Andrei Paleyes\inst{2,3}\orcidID{0000-0002-3703-8163}
	}
	\authorrunning{Ficiu, Lawrence and Paleyes}
	%
	\institute{Google (work done while at the University of Cambridge)
		\and
		Department of Computer Science and Technology, University of Cambridge, UK
		\and
		Correspondence to: \email{ap2169@cl.cam.ac.uk}
	}
	
	\maketitle              
	\begin{abstract}
		Machine learning models are deployed as a central component in decision making and policy operations with direct impact on individuals' lives. In order to act ethically and comply with government regulations, these models need to make fair decisions and protect the users' privacy. However, such requirements can come with decrease in models' performance compared to their potentially biased, privacy-leaking counterparts. Thus the trade-off between fairness, privacy and performance of ML models emerges, and practitioners need a way of quantifying this trade-off to enable deployment decisions. In this work we interpret this trade-off as a multi-objective optimization problem, and propose PFairDP, a pipeline that uses Bayesian optimization for discovery of Pareto-optimal points between fairness, privacy and utility of ML models. We show how PFairDP can be used to replicate known results that were achieved through manual constraint setting process. We further demonstrate effectiveness of PFairDP with experiments on multiple models and datasets.
		
	\end{abstract}

	\section{Introduction}
	\vspace{-10pt}
	During the past two decades, machine learning (ML) models have been integrated into a wide range of industries and applications. Crucially, these include high-stakes sectors such as healthcare \cite{ML-in-healthcare-2,ML-in-healthcare-1},
	education~\cite{ML-in-education-2,ML-in-education-3}, hiring~\cite{ML-in-hiring-1}, pretrial detention~\cite{ML-in-law-1}, financial lending~\cite{ML-in-financial-lending-2,ML-in-financial-lending-1}, and social services~\cite{ML-in-social-services-1,ML-in-social-services-2}. While significant efforts are being made towards developing more accurate ML models, practical scenarios impose additional requirements beyond performance. Protecting the users' privacy and ensuring non-discrimination against demographic subgroups are two critical prerequisites which need to be addressed prior to deploying a system in real-life settings. Consequently ML practitioners need to ensure that these systems do not put the users' data at risk and that they do not discriminate based on protected attributes before deployment. Addressing these considerations is not only a matter of moral obligation but also a legal requirement, privacy and fairness being mandated by government laws and regulations such as the General Data Protection and Regulation (GDPR) or the Equal Credit Opportunity Act (ECOA).
	
	Differential privacy~\cite{dp-main-reference-definition-from-opacus} has emerged as the de facto standard notion in privacy-preserving data analysis, enabling a rigorous and practical formalization of data privacy for applications that process sensitive information. Enforcing differential privacy to bound the risk of data disclosure involves random perturbations being applied during computations so as to limit the influence of any individual sample on the outcome of queries. Inevitably, this noise injection will also induce a loss in the overall utility of the system and determining the optimal balance between privacy and precision remains a challenging problem~\cite{trade-off-between-privacy-and-accuracy-is-challenging}. Similarly, available techniques for improving fairness in ML models rely on preprocessing the datasets used for training, modifying the learning procedure or postprocessing the final results~\cite{available-techniques-for-fairness-rely-on-pre-in-or-post-processing}. While these approaches are indeed successful in reducing bias, they also may affect the utility of the models. Finally, many works emphasise that privacy and fairness are not independent objectives~\cite{privacy-vs-fairness-1,privacy-vs-fairness-2,fair-decision-making-under-privacy-protected-data}. For example, training a model with privacy guarantees can lead to disparate performance across different groups in the population~\cite{privacy-vs-fairness-1}. Quantification of impact of differentially privacy on fairness in ML is an area of active research \cite{Arcolezi2023LocalDP,privacy-vs-fairness-2}. It is therefore necessary to consider the joint impact of utility, fairness and privacy on each other \cite{Balancing-Learning-Model-Privacy-Fairness-and-Accuracy-With-Early-Stopping-Criteria}.
	
	To address these challenges we introduce PFairDP, a pipeline for automatically quantifying the trade-off between differential privacy, fairness and performance of ML models. Compared to previous work in the field~\cite{investigating-tradeoffs-in-utility-fairness-and-dp-in-neural-networks,achieving-dp-and-fairness-in-logistic-regression}, it is not limited to training models at predefined levels of privacy, fairness or utility. Instead, our approach to this task extends DPareto proposed by Avent et al.~\cite{automatic-discovery-of-privacy-utility-pareto-fronts} that interprets the trade-off between differential privacy and utility as a choice of Pareto-optimal points in two dimensions. We incorporate fairness as an additional objective and implement a multi-objective Bayesian optimization procedure to efficiently estimate the 3D fairness-privacy-utility Pareto fronts in an automated and model-agnostic manner, relying only on empirical measurements of the model's utility, bias, and privacy. Modular structure of PFairDP supports most existing pre- and postprocessing fairness enforcement techniques. The ultimate goal of PFairDP is to efficiently determine the configurations which best balance the three-way fairness-privacy-utility trade-off. The proposed method allows decision makers to choose whether, with similar levels of utility, they want to prioritise privacy over fairness or use a configuration that balances the two. Similarly, the 3D Pareto frontiers can be used to approximate how much privacy and fairness a model can guarantee while maintaining acceptable levels of utility. The experiments performed throughout this paper illustrate the capabilities of our method and its flexibility in relation to models and datasets.
	
	\section{Motivation}
	\vspace{-10pt}
	Currently the literature investigating this three-way trade-off is limited to enforcing manually chosen levels of privacy and fairness, then evaluating the cost incurred in utility. However, due to the conflicting nature of the three objectives, we argue that this problem would be best addressed as a matter of reaching Pareto efficiency. At a Pareto-optimal solution improvement in one of the objectives necessarily deteriorates at least one of the others. Existing work relies on predefined fairness and privacy constraints and so is prevented from exploring the Pareto front. By exploring a wider range of configurations our framework provides better insight into the limitations of models and enables decision makers to prioritise certain requirements, evaluate the effects of their choices, and take informed actions before deployment.
	
	As a practical example of application of our work, consider a scenario in which census bureau is given a task to release an ML model trained on the data collected during the most recent census. The task could be small-area estimation \cite{thibaudeau2021small}, administrative lists management \cite{winkler2021cleaning}, matching of records across databases \cite{alvarez2009interstate}, classification of business entities \cite{cuffe2019using}, funds allocation \cite{abowd2019economic}. Users of the model could be another government-run department or a private contractor. As the model might be used for highly sensitive decisions, the census bureau has to take measures to make sure that citizens' data used for training is protected and that no group is discriminated against, while achieving the highest possible performance. Data scientists employed by the bureau can re-train the model as many times as necessary before releasing it to ensure this trade-off is resolved satisfactorily. Methods currently existing in the literature propose to hand-pick limits for fairness and privacy, and perform training for the highest utility possible under these constraints. This approach faces the difficult question of picking the right limits \cite{abowd2019stepping}, while also potentially missing better model's configurations. Instead, our proposed method provides the entire Pareto front between privacy, fairness and utility. This allows model developers to make informed decisions on what levels of fairness, privacy and utility can be achieved, while also facilitating a discussion with model's users on the desirable model's behaviour.
	
	\section{Related work}\label{related-work-section}
	\vspace{-10pt}
	A growing awareness of potential bias and privacy leaks of ML algorithms and datasets is reflected by increasing body of literature dedicated to exploring their effects on ML models. Feldman and Peake~\cite{end-to-end-bias-mitigation} propose an end-to-end bias mitigation framework, and demonstrate its effectiveness on the case study of bias mitigation in a deep learning setting. Morsbach et al.~\cite{architecture-matters-investigating-the-influence-of-dp-in-nn-design} investigate the relationship between neural network architectures and model accuracy under differential privacy constraints. Sharma et al.~\cite{fair-n-fair-and-robust-nn-for-structured-data} utilise the distance of data observations to the decision boundary to create a training process for neural networks that are more fair and adversarially robust, while maintaining similar level of accuracy. On a more theoretical side, Geng et al.~\cite{geng2020tight} provide tight lower and upper bounds on the privacy-utility trade-off. 
	
	While two-way privacy-utility and fairness-utility relationships are studied extensively, the literature on the joint relationship between all three metrics is scarce. Pannekoek and Spigler \cite{investigating-tradeoffs-in-utility-fairness-and-dp-in-neural-networks} and Xu et al. \cite{achieving-dp-and-fairness-in-logistic-regression} propose a constraint-based approach, where the focus is on evaluating models under predefined fairness and DP constraints. Such a constraint-based approach can leave optimal configurations unexplored and does not provide actionable information regarding the objectives' trade-offs. We show how results of both of these papers can be replicated in an automated way using our pipeline, highlighting the added value of automation over constraint-guided exploration.
	
	Further, Chester et al. \cite{3d-tradeoff-medical-data} study the same three-way relationship in the context of medical data, while Zhang et al. \cite{zhang2022trading} focus on the same relationship in federated learning context. Both papers empirically explore the trade-off, and allude to optimization techniques for answering the question ``How to choose a good balance?'', thus paving the way for our method.
	%
	
	\section{PFairDP -- Fairness- and DP-augmented pipeline}\label{fairness-and-dp-augmented-pipeline-description}
	\vspace{-10pt}
	The core component of our work is PFairDP, a Parametrized Fair and Differentially Private training pipeline for ML models. In order to provide a modular implementation that can be employed as a model-agnostic training pipeline, the architecture of PFairDP is devised into three independent modules: (1) a fairness module, (2) a DP module and (3) a training module. We implemented PFairDP using PyTorch framework~\cite{Paszke2019PyTorchAI}, Opacus \cite{opacus-original-paper}, AIF360 \cite{aif360-main-paper}, and BoTorch \cite{botorch-main-paper}\footnote{The code is openly available at \url{https://github.com/apaleyes/dp-fairness-multi-objective-bayesian-optimisation}}.
	
	At a high level, the fairness module (section \ref{fairness-module}) implements pre- and post-processing algorithms aimed at reducing bias in the datasets and the models' predictions. The DP module (section \ref{privacy-module}) augments the model, the dataset and the optimizer with DP-related capabilities required for enforcing privacy (e.g. per sample gradient computation and noise injection) and finally the training module (section \ref{training-module}) performs the training routine of the model. Given the model's final (optionally postprocessed) predictions, the pipeline returns the associated fairness measure, privacy budget, and utility, which are then used by the Bayesian optimization loop for the Pareto frontier discovery. As previously emphasised, PFairDP does not enforce predefined levels of privacy, fairness, or utility when training the models, but instead each module introduces a set of \hbox{parameters} affecting at least one of the objectives. The remainder of this section describes each module in greater detail.
	
	\subsection{Objective 1 - Fairness}
	\label{fairness-module}
	
	In the context of ML decision-making, fairness has been defined as \textit{the absence of any prejudice or favoritism toward an individual or group based on their inherent or acquired \hbox{characteristic}}~\cite{fairness-formal-definition-1,fairness-formal-definition-2}. While this formulation provides some intuition about the expected \hbox{behaviour} of an unbiased model, formalising this goal into a widely applicable and generally accepted representation remains a challenge due to the complex and multi-faceted nature of fairness, with more than 21 mathematical definitions presented in relevant literature~\cite{motivation-for-fairness-metrics-2}. 
	
	Specific fairness definitions can be separated into two main categories: \textit{individual} fairness, enforcing that similar individuals receive similar predictions, and \textit{group} fairness, which ensures that different groups (e.g. males and females) receive the same predictions with close to equal \hbox{probabilities}. The focus of this work is on the latter, as we are interested in removing bias with respect to certain protected attributes (e.g. gender, religion) and while there are a multitude of fairness metrics for this category as well, we will be using two common fairness definitions \cite{motivation-for-fairness-metrics-1,motivation-for-fairness-metrics-2} for the reminder of this paper: statistical parity difference (SPD, \cite{spd-definition}) and disparate impact (DI, \cite{disparate-impact-definition}). If needed, PFairDP can be used with other definitions. In the ideal case, the outcomes of an unbiased model which enforces group fairness would be independent of membership in a sensitive group or, in other words, the optimal values for the fairness objective are $\texttt{SPD} \sim 0$ and $\texttt{DI} \sim 1$. In practice, the fairness of a classifier will be assessed with respect to the following bounds on the two metrics introduced above:
	\begin{itemize}
		\item $0 \leq \texttt{SPD} \leq 0.1$ -- to which we will refer to as the lenient SPD threshold~\cite{lenient-threshold}.
		\item $0 \leq \texttt{SPD} \leq 0.05$ -- to which we will refer to as the strict SPD threshold~\cite{achieving-dp-and-fairness-in-logistic-regression}.
		\item $0.8 \leq \texttt{DI} \leq 1.25$ -- to which we will refer to as the standard DI threshold~\cite{disparate-impact-definition}.
	\end{itemize}
	
	PFairDP supports pre- and postprocessing techniques for enforcing fairness. Our modular implementation allows the use of any such technique, and we showcase two of them in this paper. Disparate Impact Remover (DIR, \cite{disparate-impact-definition}) is a preprocessing algorithm which modifies the values of protected attributes in order to remove distinguishing factors and improve group fairness with respect to the disparate impact metric). The procedure is parametrized with respect to one argument \textit{repair level} $\in [0, 1]$ which controls how much the distribution of the privileged and unprivileged groups should overlap, a value of 1 indicating complete overlap between the two groups. Reject Option Classification (ROC, \cite{Reject-Option-Classification}) is a postprocessing algorithm which increases fairness by assigning favorable labels to instances in the unprivileged group and unfavorable labels to instances in the privileged group in a confidence band around the decision boundary with the highest uncertainty. The ROC technique alters the outputs of a binary classifier based on its predicted posterior probabilities so that instead of solely relying on the standard decision rule $\texttt{sign}(\mathds{P}(Y = 1 | X) - \mathds{P}(Y = 0 | X))$, instances that lie close to the decision boundary are labelled based on their group membership. The algorithm is designed to minimize statistical parity difference.
	
	\subsection{Objective 2 - Privacy}
	\label{privacy-module}
	
	Differential Privacy (DP) is a mathematical framework for quantifying privacy in statistical analysis, which allows performing complex computations over large datasets while bounding the disclosure of information about individual data points~\cite{dp-main-reference-definition-from-opacus,opacus-original-paper}. To this end, DP does not define privacy as binary (has the data been exposed or not) but in terms of a ``privacy budget'', which limits the influence of any individual sample on the output of an algorithm. This notion is formally quantified by a pair of parameters $(\epsilon, \delta)$.
	
	There is no agreed-upon threshold below which algorithms are considered private~\cite{investigating-tradeoffs-in-utility-fairness-and-dp-in-neural-networks,no-agreed-upon-value-for-epsilon}. In practice~\cite{automatic-discovery-of-privacy-utility-pareto-fronts,achieving-dp-and-fairness-in-logistic-regression}, $\delta$ is chosen beforehand as a fixed small value and the privacy budget is characterised by the $\epsilon$ parameter, with smaller values indicating stronger privacy guarantees. The remainder of this work will follow this convention, with privacy reported with respect to $\epsilon$, and $\delta$ regarded as fixed: $\delta \ll \frac{1}{n}$, where $n$ is the number of records in the dataset used\footnote{This is a valid assumption as there is a proven connection between $\epsilon$ and $\delta$ \cite{balle2018privacy}}.
	
	One standard mechanism for achieving ($\epsilon$, $\delta$)-differential privacy is through the introduction of DP optimizers in the training procedure~\cite{dp-optimizers-for-training-neural-networks}. The DP-aware optimization used in PFairDP adds noise to the parameter gradients in every iteration, thus preventing training samples from being memorized and altering the outputs of the model. The amount of noise added during computations is controlled by two parameters: a \textit{noise multiplier} representing the amount of noise added to the average of the gradients in a batch, and a \textit{clipping norm} representing the maximum $l_{2}$ norm of per-sample gradients, which bounds the impact of a single sample on the model.
	
	\subsection{Objective 3 - Utility}
	\label{training-module}
	Finally, the training module of PFairDP implements a standard training procedure for ML models. In the case of neural networks, which is the focus of experimentation section of our paper, this introduces three additional parameters: the \textit{batch size}, the \textit{number of epochs}, and the optimizer's \textit{learning rate}. These hyperparameters affect not only the final utility of the model, but also the privacy budget (e.g. a larger batch size increases the privacy budget) and the fairness level (e.g. training for longer on a debiased dataset will increase the fairness level while decreasing the privacy budget).
	
	In this work we focus on binary classification problems so that fairness can be quantified with respect to whether (un)favorable outcomes are assigned to individuals in the privileged and unprivileged groups. If necessary, it is possible to formulate a generalization beyond this use case. 

	\subsection{Determining optimal configurations}
	\label{determining-optimal-configurations}
	
	PFairDP can be seen as a black-box function which, given a set of hyperparameters, returns the associated values for the three objectives. Proceeding with this mental model, the task of determining configurations which best balance the three-way fairness-privacy-utility trade-off becomes a multi-objective optimization problem. Multi-objective Bayesian optimization (MOBO) provides a computationally effective way to address this task.
	
	Bayesian Optimization (BO)~\cite{bayesian-optimisation-original-paper} is a class of sample-efficient optimization techniques which have shown success in the optimization of black-box objective functions with high evaluation costs, establishing a new state-of-the-art in ML hyperparameter tuning~\cite{bayes-opt-is-state-of-the-art-in-hyperparameter-tuning} as well as various other domains~\cite{bayes-opt-application-1,bayes-opt-application-2}. At a high level, given an expensive black-box function $f$ that needs to be minimized, BO learns a probabilistic \textit{surrogate model} of the function based on a limited set of evaluations, and relies on an \textit{acquisition function} to determine the next point to be evaluated based on the learned probabilistic model. The main advantage of the BO framework is that while the true function $f$ might be expensive to evaluate, the surrogate-based acquisition function is not and can therefore be used to determine a set of potential candidates that minimize the objective. Because the main focus of this paper is in simultaneously optimizing for multiple objectives (the utility of the model, the privacy budget, and the fairness level) we introduce below the formalism of the multi-objective BO.
	
	\textbf{Pareto Fronts.} For multi-objective optimization problems there is usually no single solution that can simultaneously optimize all objectives; rather, the goal is to identify the set of \textit{Pareto optimal} solutions such that improving any one of the objectives comes at the cost of deteriorating another. Formally (assuming a minimization problem), we say that a solution $\textbf{f}(x)$ \textit{Pareto dominates} another solution $\textbf{f}(x')$ if for all $k \in \{1, 2, ..., m\}$ it holds that $f^{k}(x) \leq f^{k}(x')$ and there exists at least one $k$ which satisfies $f^{k}(x) < f^{k}(x')$, where $m$ is the number of objectives.
	The \textit{Pareto frontier} of optimal trade-offs is defined as the set of non-dominated solutions and the goal of a multi-objective optimization algorithm is to determine an approximate Pareto frontier. 
	
	\textbf{Multi-objective Bayesian Optimization.} Without loss of generality, we consider a minimization problem of $m \geq 1$ objective functions $f_{1}: \mathbf{X} \rightarrow \mathbb{R}$, ..., $f_{m}: \mathbf{X} \rightarrow \mathbb{R}$, where $\mathbf{X} \subset \mathbb{R}^{n}$ is a bounded set. If we assume $\mathbf{D} = (x_{i}, \textbf{f}(x_{i}))_{i = 1}^{k}$ to be a dataset of known evaluations of the objectives\footnote{In the context of our experiment $\mathbf{D}$ would represent the associated privacy budget, fairness level, and utility for a set of hyperparameters configurations, determined from previous evaluations of the pipeline.}, the MOBO procedure can be defined as a four-step process repeated for a predefined number of iterations:
	\begin{enumerate}
		\item Fit a surrogate model of the objectives using the observed data $\mathbf{D}$. In our work we use Gaussian processes (GP, \cite{gp-main-citation}).
		\item Determine the posterior distribution $\mathds{P}(\textbf{f} | \mathbf{D})$ over the true function values $\textbf{f}$ using the surrogate model.
		\item Collect the next evaluation point $x_{k + 1}$ at the estimated global maximum of the acquisition function, based on the observed values $\mathbf{D}$ and the posterior. We use the expected hypervolume improvement (EHVI)~\cite{hypervolume-improvement} as the acquisition function.
		\item Update the set of observations $\mathbf{D}$ with the new sample.
	\end{enumerate}
	
	\textbf{EHVI Acquisition Function.} Acquisition function is a key component of the MOBO procedure that determines the new evaluation point while balancing exploration and exploitation. PFairDP uses expected hypervolume improvement (EHVI) acquisition function, which we now explain.
	
	Given a set $\mathbf{P} \subset \mathbb{R}^{m}$, the \textit{hypervolume} of $\mathbf{P}$ is defined as the hypervolume of its dominated region bounded by a fixed user-defined reference point. The reference, or anti-ideal, point is a point in the objective space dominated by all of the Pareto-optimal solutions and is usually chosen by the practitioner based on domain knowledge. For our use case, this could be selected to indicate the worst possible values for the utility of the model, the fairness level and the privacy budget. Formally, this can be expressed as $\mathcal{H}(\mathbf{P}) = \mu(\{\mathbf{y} \in \mathbf{R}^{m} \: | \: \mathbf{y} 	\prec \mathbf{r} \: \mathrm{and} \: \exists \; p \in \mathbf{P}: p \prec \mathbf{y} \})$, where $\mu$ denotes the standard Lebesgue measure on $\mathbb{R}^{m}$. A larger hypervolume implies that the points in $\mathbf{P}$ are closer to the true (unknown) Pareto front.
	
	We now define the \textit{hypervolume improvement} of a vector $\mathbf{y} \in \mathbb{R}^{m}$ with respect to a set $\mathbf{P}$ as $\mathcal{H}_{I}(\mathbf{y}, \mathbf{P}) = \mathcal{H}(\mathbf{P} 	\cup \{ \mathbf{y} \}) - \mathcal{H}(\mathbf{P})$, which is positive only if the point $\mathbf{y}$ lies in the set of points non-dominated by $\mathbf{P}$. In the context of MOBO, $\mathbf{P}$ would be the set of samples evaluated so far and $\mathbf{y}$ the random output of the GPs used for modelling the objectives. Finally, the expected hypervolume improvement is defined as the expected value of $\mathcal{H}_{I}(\mathbf{y}, \mathbf{P})$ over the distribution $\mathds{P}(\mathbf{y})$, $E\mathcal{H}_{I}(\mathbf{y}) = \mathds{E}[\mathcal{H}_{I}(\mathbf{y}, \mathbf{P})]$. In other words, EHVI computes the expected gain in hypervolume of observing one candidate point. 
	
	In the MOBO procedure we therefore choose the point which maximizes the EHVI as the new sample to evaluate. The optimization loop is allowed to execute within a predefined budget of evaluations. When the loop terminates, the Pareto front is approximated based on the set of non-dominated solutions determined throughout the sampling process. The quality of the Pareto front is evaluated with respect to the hypervolume indicator, which calculates the volume encapsulated between the reference point and the Pareto-optimal points.
	
	\section{Automation of existing approaches}\label{automation-of-existing-approaches}
	\vspace{-10pt}
	The modularity of PFairDP together with the high level abstractions introduced in each module enable a wide range of bias mitigation algorithms and DP configurations to be integrated into the pipeline. In this section we show how our framework can be used to reproduce experiments from the literature in a more automated manner. We show that with PFairDP these constraint-based approaches become special cases of our method whose results can be reproduced with only minor modifications of our pipeline. 
	
	\subsection{Trade-offs in Utility, Fairness and DP in Neural Networks}
	\label{automated-experiment-1}
	The work of Pannekoek and Spigler \cite{investigating-tradeoffs-in-utility-fairness-and-dp-in-neural-networks} in enabling an ethical and legal use of ML algorithms most closely resembles the aims of our project. More specifically, while the authors also evaluate the privacy-utility-fairness trade-off in ML, their proposed framework focuses on evaluating the models under predefined fairness and DP constraints. The authors evaluate four models: a Simple (S-NN), a Fair (F-NN), a Differentially Private (DP-NN), and a Differentially Private and Fair Neural Network (DPF-NN), detailed description of which can be found in Appendix \ref{appendix-Pannekoek-models}. They use Adult dataset (described in Appendix \ref{appendix-datasets}) and regard `sex' as the sensitive attribute (treated as binary). In terms of data preprocessing, they apply following transformations: list-wise deletions, one-hot encoding of categorical variables, normalization of continuous variables, train-dev-test splits in proportions of 53.4\%, 13.3\%, 33.3\% of the total. To enable a fair comparison, we also replicate these preprocessing steps in our experiment.
	
	The authors train each of the four models on the Adult dataset and evaluate utility in terms of the mean accuracy of the models and fairness with respect to the average risk difference after ten independent runs. The privacy level is regarded as a fixed constraint ($\epsilon$ = 0.1 for $\delta$ = 0.00001). As shown in table \ref{table:replicating-trade-offs-in-fairness-dp-and-accuracy}, replicating these experiments with PFairDP is a matter of only enabling certain modules in the pipeline and configuring the core module with the same settings. The appropriate level of noise to be added in the DP module is identified beforehand based on the required privacy level $\epsilon$, the sampling rate, the number of epochs and the fixed $\delta$. In the postprocessing module the ROC algorithm is initialised with the default parameters for consistency with original work. 
	
	\begin{table}[t]
		\caption{Models defined by Pannekoek and Spigler and their PFairDP equivalent configurations.}
		\begin{center}
			\begin{tabular}{ c|c|c|c| }
				\cline{2-4}
				~ & \multicolumn{3}{|c|}{\textbf{PFairDP configuration}} \\\hline
				\multicolumn{1}{|c|}{\textbf{Original model}} & \textbf{Preprocessing} & \textbf{DP} & \textbf{Postprocessing} \\\hline
				\multicolumn{1}{|c|}{S-NN} & Disabled & Disabled & Disabled \\\hline
				\multicolumn{1}{|c|}{F-NN} & Disabled & Disabled & ROC \\\hline
				\multicolumn{1}{|c|}{DP-NN} & Disabled & Fixed noise level & Disabled \\\hline
				\multicolumn{1}{|c|}{DPF-NN} & Disabled & Fixed noise level & ROC \\\hline
			\end{tabular}
		\end{center}
		\label{table:replicating-trade-offs-in-fairness-dp-and-accuracy}
		\vspace{-20pt}
	\end{table}
	
	The performance of the four original models and their PFairDP counterparts is displayed in table \ref{table:results-pannekoek-and-spigler}. 
	As expected, models implemented with PFairDP exhibit most of the patterns observed by the authors: (1) non-fair models rank best in terms of accuracy but have the highest bias; (2) the fair, non-private model reduces bias below what is considered to be the standard lenient threshold~\cite{lenient-threshold} but also performs worse in terms of utility; (3) the DP and fair model is the most effective in mitigating bias, reducing the risk difference below the standard strict threshold~\cite{achieving-dp-and-fairness-in-logistic-regression}. One discrepancy noted during our evaluations arose in the comparison of the differentially private and fair DPF-NN model and the fair-only F-NN model. While Pannekoek and Spigler note that DPF-NN outperforms F-NN with respect to both the fairness and utility, our evaluations suggest that the latter is in fact negatively affected. This result is consistent with existing literature \cite{discrepancy-1,discrepancy-2,discrepancy-3}, which notes that imposing DP constraints on neural networks leads to decreases in the utility of the models.
	
	\begin{table*}[t]
		\caption{Accuracy and risk difference of models proposed by Pannekoek and Spigler \cite{investigating-tradeoffs-in-utility-fairness-and-dp-in-neural-networks} and their PFairDP equivalents. The performance of PFairDP closely matches that of the original models when one or no module is enabled. However, when both the DP and fairness modules are combined to replicate the DPF-NN model we note that accuracy continues to decrease. This result is consistent with existing literature, but contradicts the observations by Pannekoek and Spigler.}
		\begin{center}
			\begin{tabular}{c|c|c|c|c| } 
				\cline{2-5}
				& \multicolumn{2}{c|}{\textbf{Pannekoek and Spigler}} & \multicolumn{2}{c|}{\textbf{PFairDP}}  
				\\\cline{2-5}
				& \textbf{Accuracy} & \textbf{Risk Difference} & \textbf{Accuracy} & \textbf{Risk Difference} \\\hline
				\multicolumn{1}{|c|}{S-NN} & 84.14 ± 0.34 & 0.1310 ± 0.0147 & 85.03 ± 0.09 & 0.1800 ± 0.0039 \\\hline
				\multicolumn{1}{|c|}{DP-NN} & 84.03 ± 0.05 & 0.1355 ± 0.0024 & 82.58 ± 0.28 & 0.1394 ± 0.0192 \\\hline
				\multicolumn{1}{|c|}{F-NN} & 79.25 ± 3.50 & 0.0566 ± 0.0065 & 78.48 ± 1.04 & 0.0294 ± 0.0033 \\\hline
				\multicolumn{1}{|c|}{DPF-NN} & 82.98 ± 0.19 & 0.0475 ± 0.0020 & 74.16 ± 2.63 & 0.0270 ± 0.0085 \\\hline
			\end{tabular}
		\end{center}
		\label{table:results-pannekoek-and-spigler}
		\vspace{-10pt}
	\end{table*}
	
	
	\subsection{Achieving DP and Fairness in Logistic Regression}
	\label{automated-experiment-2}
	Xu et al.~\cite{achieving-dp-and-fairness-in-logistic-regression} propose a privacy-preserving and fair logistic regression (LR) model which incorporates a penalising term to the objective in order to ensure fairness and a functional mechanism \cite{functional-mechanism} which also perturbs the objective function to enforce DP. While the framework proposed by Xu et al. is indeed successful in achieving both fairness and DP for a logistic regression model, we illustrate below how PFairDP can be used to achieve comparable results without relying on model specific augmentation procedures.

	
	
	The authors propose a total of four modified versions of logistic regression (LR): PrivLR, FairLR, PFLR, and PFLR* (the latter provides stronger fairness guarantees compared to PFLR). Their detailed description can be found in Appendix \ref{appendix-Xu-models}. Risk difference is used as the fairness metric, and evaluation is done with the Adult dataset. PFairDP configurations for this experiment are summarised in table \ref{table:replicating-logistic-regression-configurations}. All of the models implemented in PFairDP are trained for a fixed duration of 100 epochs using the Adam optimizer with minibatches of size 20 and a fixed learning rate of 1e-3. One important distinction between our work and that of Xu et al. is that their definition of DP assumes $\delta = 0$ (known in the literature as $\epsilon$ - differential privacy~\cite{dp-main-reference-definition-from-opacus}) which is incompatible with our Opacus implementation for the privacy module. As such, in all of our evaluations $\delta$ is set to be a sufficiently small and fixed constant ($\delta = 0.00001$, as used in~\cite{investigating-tradeoffs-in-utility-fairness-and-dp-in-neural-networks} on the same task). 
	
	\begin{table}[t]
		\caption{Models defined by Xu et al.~\cite{achieving-dp-and-fairness-in-logistic-regression} and PFairDP equivalents which achieve comparable performance.}
		\begin{center}
			\begin{tabular}{ c|c|c|c| } 
				\cline{2-4}
				~ & \multicolumn{3}{c|}{\textbf{PFairDP configuration}} \\\hline
				\multicolumn{1}{|c|}{\textbf{Original model}} & \textbf{Preprocessing} & \textbf{DP} & \textbf{Postprocessing} \\\hline
				\multicolumn{1}{|c|}{PrivLR} & Disabled & Fixed noise level & Disabled \\\hline
				\multicolumn{1}{|c|}{FairLR} & DIR & Disabled & Disabled \\\hline
				\multicolumn{1}{|c|}{PFLR} & DIR & Fixed noise level & Disabled \\\hline
				\multicolumn{1}{|c|}{PFLR*} & DIR & Fixed noise level & ROC \\\hline
			\end{tabular}
		\end{center}
		\label{table:replicating-logistic-regression-configurations}
		\vspace{-10pt}
	\end{table} 
	
	Table \ref{table:replicating-privlr} displays the performance of the three original models and their PFairDP counterparts with respect to the utility and fairness objectives for various privacy constraints. Note that results for FairLR are not displayed because Xu et al.~\cite{achieving-dp-and-fairness-in-logistic-regression} only include evaluations of PrivLR, PFLR, and PFLR* when varying the privacy budget.
	
	\begin{table*}[t]
		\caption{Accuracy and risk difference of models proposed by Xu et al. for different privacy budgets $\epsilon$ and comparison with their PFairDP equivalents. When no postprocessing is applied to the models' predictions, PFairDP provides similar fairness guarantees to PrivLR and PFLR while achieving significantly higher levels of utility. Integrating the postprocessing procedure further reduces bias to levels similar to PFLR* but also degrades the accuracy of the model.}
		\begin{center}
			\begin{tabular}{|c|c|c|c|c|c| } 
				\hline
				& \multirow{2}*{$\epsilon$} & \multicolumn{2}{c|}{\textbf{Xu et al.}} & \multicolumn{2}{c|}{\textbf{PFairDP}}  
				\\\cline{3-6}
				& & \textbf{Accuracy} & \textbf{Risk Difference} & \textbf{Accuracy} & \textbf{Risk Difference} \\\cline{1-6}
				& 0.1 & 0.6263 ± 0.1480 & 0.0883 ± 0.0805 & 0.7845 ± 0.0022 & 0.0271 ± 0.0041 \\\cline{2-6}
				PrivLR & 1 & 0.7238 ± 0.0612 & 0.0502 ± 0.0581 & 0.8027 ± 0.0006 & 0.0699 ± 0.0019 \\\cline{2-6}
				& 10 & 0.7270 ± 0.0877 & 0.1459 ± 0.0798 & 0.8073 ± 0.0004 & 0.0762 ± 0.0026 \\\cline{1-6} 
				& 0.1 & 0.6172 ± 0.1187 & 0.0351 ± 0.0493 & 0.7749 ± 0.0029 & 0.0126 ± 0.0026 \\\cline{2-6}
				PFLR & 1 & 0.7400 ± 0.0182 & 0.0213 ± 0.0258 & 0.7981 ± 0.0008 & 0.0321 ± 0.0020 \\\cline{2-6}
				& 10 & 0.7631 ± 0.0155 & 0.0338 ± 0.0255 & 0.8024 ± 0.0004 & 0.0339 ± 0.0009 \\\cline{1-6} 
				& 0.1 & 0.7491 ± 0.0040 & 0.0028 ± 0.0039 & 0.7273 ± 0.0905 & 0.0087 ± 0.0059 \\\cline{2-6}
				PFLR* & 1 & 0.7552 ± 0.0092 & 0.0053 ± 0.0070 & 0.7526 ± 0.0647 & 0.0098 ± 0.0065 \\\cline{2-6}
				& 10 & 0.7632 ± 0.0093 & 0.0204 ± 0.0140 & 0.7557 ± 0.0544 & 0.0153 ± 0.0084 \\\cline{1-6}
			\end{tabular}
		\end{center}
		\label{table:replicating-privlr}
		\vspace{-10pt}
	\end{table*}
	
	
	First considering the models which only enforce DP and no debiasing procedures (PrivLR and PFairDP with only the DP module enabled), we note that the degradation in utility is significantly less pronounced in PFairDP from the original non-private and non-fair logistic regression model, which achieves a mean accuracy of 84\%. However, this improvement does not come at the cost of fairness, with risk difference generally being lower in PFairDP and below the lenient SPD threshold. Together with the observation that the fairness of the model decreases as the privacy level is decreased, this suggests that the noise injected during the training procedure of the neural network acts as a fairness regularising term, perturbing the final predictions to include more positive predictions for individuals in the unprivileged group and reducing the bias of the model. 
	
	A similar pattern can be observed when comparing PFLR and PFairDP with DP and fairness preprocessing enabled, both models achieving similar performance in terms of risk difference but latter showing higher accuracy. Additionally, significant decreases can be noted in terms of variance for both objectives (by a factor of 10 on average), suggesting that our training pipeline is much more stable than the objective perturbation approach employed by the authors. As expected, increasing the value of $\epsilon$ (i.e. reducing the privacy level by decreasing the amount of noise) improves the utility of both PFLR and PFairDP-PFLR. 
	
	Finally, we note that while combining the DIR fairness pre-processing and ROC post-processing procedures leads to a decrease in accuracy compared to PFairDP-PFLR, the performance of our model closely matches that of PFLR*. When the privacy constraint is relaxed, PFairDP only exhibits minor degradation with respect to the fairness objective whereas the risk difference of the logistic regression model increases by a factor of 4 when the privacy budget varies from 0.1 to 10. 
	
	These results suggest that PFairDP can enable an automated way of generating fair and differentially private models without relying on model-specific augmentations and constraint-setting.
	
	\section{Automatic Discovery of Fairness - Privacy - Utility Pareto Fronts}\label{new-experiments-fairness-privacy-utility-pareto-fronts}
	\vspace{-10pt}
	This section demonstrates the effectiveness of PFairDP for determining Pareto fronts on two binary classification tasks. Here PFairDP is evaluated in its intended execution mode, with all the parameters in the fairness, DP, and training modules varying within predefined ranges.
	
	\subsection{Experimental Setup}
	\hspace{\parindent}\textbf{Datasets.} Experiments are performed on two standard benchmark datasets in the fair and private machine learning literature: Adult and MEPS. See Appendix \ref{appendix-datasets} for their general descriptions and motivation for using bias mitigation and privacy preserving techniques on these datasets. 
	
	\textbf{Models.} The model trained on the Adult dataset follows the same architecture used by Pannekoek and Spigler \cite{investigating-tradeoffs-in-utility-fairness-and-dp-in-neural-networks} (previously described in section \ref{automated-experiment-1}). For MEPS we use a similar configuration suggested by Sharma et al.~\cite{fair-n-fair-and-robust-nn-for-structured-data}, in the form of a two layer dense neural network with 30 hidden units in each layer and ReLU activations, followed by a sigmoid on the output layer.
	
	\textbf{PFairDP Module Configurations and Optimization Domains.} 
	Both training datasets encode some level of favoritism towards the privileged group above the acceptable threshold (see Appendix \ref{appendix-datasets} for details). To reduce the level of disparate impact potentially exhibited by models trained on these datasets, the fairness module is enabled with DIR preprocessing for varying values of the \textit{repair level} hyperparameter. In both experiments the DP module is enabled using its standard parameters: the \textit{noise multiplier}, controlling the amount of noise added to the average of the gradients in a batch, and the \textit{clipping norm}, which bounds the maximum $l_{2}$ norm of per-sample gradients. In order to also explore the effect of different optimization algorithms used in the training procedure, experiments are performed using privatized versions of the Adam optimizer~\cite{ADAM-reference} for Adult and stochastic gradient descent (SGD)~\cite{SGD-reference} for MEPS. The same three hyperparameters are varied in both experiments, namely the \textit{number of training epochs}, the optimizer's \textit{learning rate} and the \textit{batch size}. Additionally, we have applied transformations to the PFairDP outputs for convenience of GP modelling. These transformations, as well as value ranges for all input parameters, can be found in Appendix \ref{appendix-domains}.
	
	\textbf{Anti-ideal point selection.} Previously introduced when describing the implementation of the MOBO procedure in \hbox{section} \ref{determining-optimal-configurations}, an anti-ideal point is required for computing the hypervolume improvements of the Pareto frontiers determined during the optimization loops and for evaluating the quality of the frontiers. The anti-ideal point is dominated by all the Pareto-optimal solutions and defines bounds on the worst possible values for each of the three objectives. In our experiments we use $(\textrm{acc}, \textrm{fair}, \textrm{DP}) = (0, 1, 1)$ as the anti-ideal point\footnote{Note that the transformations described above are also applied to the reference point.}, encoding our interest in Pareto frontiers which capture a practical privacy range ($\epsilon \leq 1$) across all possible utility and fairness values (since these can never exceed 1 or be below 0).
	
	\subsection{Experimental Results}
	This section evaluates the performance of the MOBO procedure in approximating Pareto frontiers on the two binary classification tasks (Adult and MEPS). Additionally, two commonly used hyperparameter tuning strategies, grid and random search, are used as baselines for comparison. For each task we execute:
	\begin{itemize}
		\item 250 BO iterations with 16 random initial configurations.
		\item 256 rounds of grid search, using 4 uniform samples per parameter, with fixed batch size and the learning rate.
		\item 300 rounds of random search.
	\end{itemize}
	
	Sparse 3D Pareto fronts produced by MOBO procedure are hard to visualise in a 2D image, therefore we refrain from showing these plots here. Our implementation produces interactive 3D visualisations of the Pareto front that can be inspected and used for decision making interactively. For further discussion please refer to Appendix \ref{appendix:3d-pareto-vis}. However, decision makers can make more informed decisions when balancing the three-way trade-off even without a complete plot of the frontier. For example, as shown in table \ref{table:same-accuracy-different-fairness-and-privacy}, similar accuracies can be obtained with different effects on privacy and fairness; policy makers can decide which one they want to prioritise.
	
	\begin{table}[t]
		\caption{Configurations which achieve similar levels of utility while providing different privacy and fairness guarantees. PFairDP provides flexibility to policy makers in deciding whether they want to prioritize one objective over the other.}
		\begin{center}
			\begin{tabular}{ |c|c|c|c| } 
				\hline
				\textbf{Dataset} & \textbf{Accuracy} & \textbf{Privacy level ($\epsilon$)} & \textbf{Fairness level (SPD)} \\\hline
				Adult & 80.75\% & 0.3899 & 0.0515 \\\hline
				Adult & 80.50\% & 0.0645 & 0.0552 \\\hline
				MEPS & 84.46\% & 0.4424 & 0.0161 \\\hline
				MEPS & 84.72\% & 0.2546 & 0.0178 \\\hline
			\end{tabular}
		\end{center}
		\label{table:same-accuracy-different-fairness-and-privacy}
		\vspace{-10pt}
	\end{table}
	
\begin{figure*}[t]
		\centering
		\includegraphics[width=0.95\textwidth]{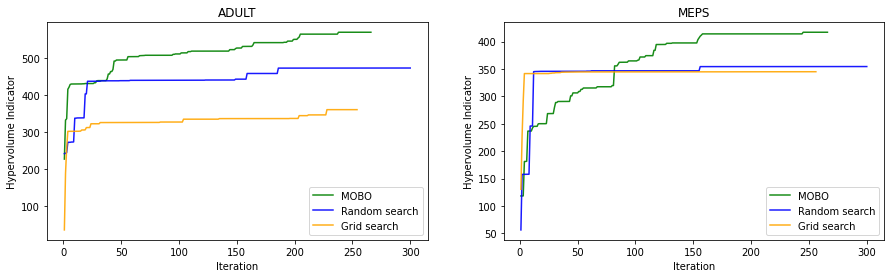}
		\caption[Hypervolume improvements of the Pareto frontiers as new configurations are sampled on Adult and MEPS]{Hypervolume improvements as new configurations are sampled on the Adult and MEPS datasets. Even though both the random and grid search procedures are allowed to explore more of the hyperparameter space, the hypervolumes of the Pareto frontiers approximated by MOBO are higher, indicating that it is better able to characterize the true Pareto fronts.}
		\label{fig:hypervolume_improvements_combined}
		\vspace{-10pt}
	\end{figure*}

	In order to compare the performance of the three sampling procedures in approximating the true Pareto frontiers, we explore their improvements with respect to the hypervolume indicator as new configurations are evaluated. As illustrated in figure \ref{fig:hypervolume_improvements_combined} the MOBO procedure generates Pareto frontiers with higher hypervolumes in both experiments, compared to random and grid search sampling strategies\footnote{Additionally, we observe that random search consistently outperforms grid search, an effect well known in the literature \cite{random-search-better-than-grid-search}.}. These improvements in performance can be observed even when only about 100 configurations have been sampled and hold in spite of the fact that the other two methods are allowed to explore for more iterations, which emphasises the increased sample efficiency of the MOBO procedure.
	
	It is important to discuss computational efficiency of PFairDP. While the MOBO procedure is able to find better Pareto frontiers, its optimization loop has sequential nature. In contrast, the grid and random search implementations can be easily parallelized to increase the number of samples evaluated within the same time frame. Nevertheless, when hardware limitations are a constraint our MOBO implementation enables an efficient procedure for determining optimal operating points for a given model. Furthermore, PFairDP can be improved by leveraging existing batch MOBO methods \cite{paleyes2022hippo}.
	
	\section{Conclusions}\label{conclusion-and-discussion}
	\vspace{-10pt}
	The premise put forward by this paper is that due to the conflicting nature of the fairness, privacy and utility, a shift in perspective is required in order to optimally balance the three-way trade-off. Currently prevalent manual constraint setting approach leads to discovery of only a subset of the optimal solutions. As an alternative, we introduced PFairDP, an automatic, efficient and holistic approach for training differentially private and fair ML models that reaches empirical Pareto-optimality with respect to the three objectives using multi-objective Bayesian optimization. We showed that PFairDP can be used to automate previous research in the privacy- and fairness-aware ML. We also evaluated our method on classification tasks with multiple bias mitigation methods, models, datasets and optimizers in order to showcase the modularity and versatility of our implementation, as well as its ability to provide actionable information to policy makers tasked with balancing the privacy-fairness-utility trade-offs of a ML system before deployment.

\bibliographystyle{splncs04}
\bibliography{bibliography}
	
\appendix

\clearpage
\noindent{\huge\bfseries APPENDICES\par}

\section{Models used in ``Trade-offs in Utility, Fairness and DP in Neural Networks''}\label{appendix-Pannekoek-models}

This section describes the models used by Pannekoek and Spigler \cite{investigating-tradeoffs-in-utility-fairness-and-dp-in-neural-networks}. The authors evaluate four models: a Simple (S-NN), a Fair (F-NN), a Differentially Private (DP-NN), and a Differentially Private and Fair Neural Network (DPF-NN).

The Simple Neural Network (S-NN) consists of three fully connected layers with six neurons in the first and second layer and one neuron the final layer. The ReLU activation is used for the first two layers and sigmoid for the final one. Training is performed for a fixed duration of 20 epochs using the Adam optimizer and binary cross-entropy as the loss function.

The architecture of the Fair Neural Network (F-NN) is identical to that of the S-NN, the only difference being that the ROC postprocessing algorithm is used to reduce bias in the final predictions of the model. Similarly, the architecture of the Differentially Private Neural Network (DP-NN) is identical to that of the S-NN, except that a differentially private variant of the original optimizer is used for training in order to add noise to the gradients and ensure DP. Finally, the Differentially Private and Fair Neural Network (DPF-NN) integrates both the postprocessing step and the DP optimizer.

\section{Models used in ``Achieving DP and Fairness in Logistic Regression''}\label{appendix-Xu-models}

This section describes the models used by Xu et al.~\cite{achieving-dp-and-fairness-in-logistic-regression}. The authors propose a total of four modified versions of logistic regression (LR): PrivLR, FairLR, PFLR, and PFLR*.

The core architecture is PFLR, a modified version of LR which enforces DP by injecting Laplacian noise into the polynomial coefficients of the original objective function and fairness by including an additional penalty which aims to minimize the overall discrimination of the model (quantified with respect to the risk difference metric).

PrivLR and FairLR are single objective versions of PFLR: the former enforces DP using the functional mechanism whereas the latter only includes the fairness penalty term. Finally, PFLR* is introduced as an enhanced version of the core model which, instead of enforcing the fairness objective through a separate penalty term, incorporates it into the DP functional mechanism, leading to a reduction in the overall noise levels while preserving the privacy budget and the fairness of the model. 

\section{Datasets}\label{appendix-datasets}

Here we give general description of the datasets used in our work.

The Adult dataset \cite{adult-dataset} has been particularly relevant in research on privacy and fairness due to the presence of personally identifiable information (PII) and sensitive attributes which can potentially be used for identifying individuals or induce bias in models trained on it. The dataset contains 45,222 records and 14 demographic features, with the task of predicting whether the income of a person is below or above 50,000 USD.

The Medical Expenditure Panel Survey (MEPS, \cite{MEPS-dataset}) is a dataset pertaining to the healthcare domain, produced by the US Department of Health and Human Services and assumed to be representative of people's healthcare expenditures in the US. The MEPS dataset includes multiple sensitive attributes, along with other non-protected attributes such as health services used, costs and frequency of services, the classification task being that of predicting whether a person would have high utilization of medical services (defined as requiring at least 10 trips for some sort of medical care).

Table \ref{table:datasets-used-in-pfair-dp} summarises the most relevant aspects for each of the two datasets as well as the level of disparate impact and statistical parity difference, indicating the extent to which bias is inherently embedded in the training data. For both Adult and MEPS the DI level is below the standard acceptable threshold of 0.8 and the SPD is above the lenient threshold of 0.1, which motivates bias usage of mitigation techniques. Privacy is a requirement in this context as well because both datasests include sensitive information for individuals (financial and health related data). Therefore, the two binary classification problems described above are relevant for evaluating the performance of our proposed framework in optimally enforcing fairness and differential privacy.

\begin{table}[ht]
	\caption[Summary of the datasets used in our experiments]{Summary of the datasets used in our experiments. Privacy must be enforced when training models on this data as it contains sensitive financial and health related information. Bias mitigation is also necessary because both datasets embed some level of unfairness towards the unprivileged groups (indicated by DI and SPD outside the acceptable thresholds). }
	\begin{center}
		\begin{tabular}{ |c|c|c| }
			\hline
			\textbf{Dataset} & Adult & MEPS \\\hline
			\textbf{Training Size} & 39073 & 12540 \\\hline
			\textbf{Test Size} & 9769 & 3135 \\\hline
			\makecell{\textbf{Protected} \\ \textbf{Attribute}} & Sex & Race \\\hline
			\makecell{\textbf{Favorable} \\ \textbf{Label}} & Income $>$ $\$5000$ & Utilization $\geq 10$ \\\hline
			\makecell{\textbf{Disparate} \\ \textbf{Impact}} & 0.37 & 0.49 \\\hline
			\textbf{SPD} & 0.19 & 0.13\\\hline
		\end{tabular}
	\end{center}
	\label{table:datasets-used-in-pfair-dp}
\end{table}

\clearpage

\section{PFairDP input and output domains}\label{appendix-domains}

This section gives some additional details on the input and output domains used in the experimentation section of the paper.

Table \ref{table:optimisation-domains}  displays the optimization domains used in our MOBO experiments. Some of the random sampling distributions proposed by Avent et al.~\cite{automatic-discovery-of-privacy-utility-pareto-fronts} are also employed in our random sampling procedure, as these have been shown to have a positive effect on performance over naive uniform sampling and improve the quality of the Pareto frontiers.

\begin{table}[h]
	\caption[Optimization domains used in each of the two experimental settings]{Optimization domains used in each of the two experimental settings. These ranges have been chosen by reviewing relevant literature~\cite{aif360-main-paper,opacus-original-paper,automatic-discovery-of-privacy-utility-pareto-fronts} for the debiasing procedures and DP mechanisms used.}
	\begin{center}
		\begin{tabular}{ |c|c|c|c| }
			\hline
			& \textbf{Dataset} & Adult & MEPS \\\hline
			\textbf{Fairness} & \makecell{\textbf{Repair} \\ \textbf{level}} & [0, 1] & [0, 1] \\\hline
			\multirow{2}{*}{\textbf{DP}} & \makecell{\textbf{Noise} \\ \textbf{multiplier}} & [1.0, 5.0] & [1.0, 5.0] \\
			\cline{2-4}
			& \makecell{\textbf{Clipping} \\ \textbf{norm}} & [0.1, 2] & [0.1, 2] \\\hline
			\multirow{3}{*}{\textbf{Training}} & \makecell{\textbf{Number of} \\ \textbf{epochs}} & [30, 128] & [30, 128] \\
			\cline{2-4}
			& \makecell{\textbf{Learning} \\ \textbf{rate}} & [1e-3, 0.1] & [1e-3, 0.1] \\
			\cline{2-4}
			& \makecell{\textbf{Batch} \\ \textbf{size}} & [16, 64] & [16, 64]\\\hline
		\end{tabular}
	\end{center}
	\label{table:optimisation-domains}
\end{table}

The goal of our experiments is to determine three-dimensional Pareto frontiers for the three objectives of interest, namely privacy level (evaluated with respect to the privacy budget $\epsilon$), utility (here the accuracy of the models), and fairness (quantified as the statistical parity difference exhibited by the model). However, the output domains of these three metrics may not be well modeled by GPs, which model outputs on the entire real line. To address this, we follow the approach suggested by Avent et al.~\cite{automatic-discovery-of-privacy-utility-pareto-fronts} and transform the outputs as illustrated in table~\ref{table:transformed-objectives}. Furthermore, we choose these transformations so that the optimization task can be treated as a maximization problem across all dimensions (previously $\epsilon$ and SPD needed to be minimized, whereas accuracy had to be maximized). Importantly, should the users want to replace fairness metrics used by PFairDP, they shall provide similar transformation for it\footnote{For example, one valid transformation that can be applied for replacing disparate impact as the fairness optimization objective is $\log(\frac{x}{2}) + \log(1 - \frac{x}{2})$.}.

\begin{table}[th]
	\caption[Original output domains for each of the three objectives and transformations applied for the MOBO procedure]{Original output domains for each of the three objectives and transformations applied for the MOBO procedure. The proposed functions transform the original output domains to the real axis so that they can be better modeled using Gaussian processes. A secondary purpose is to also convert the optimization tasks into maximisation across all objectives, a requirement imposed by the Bayesian optimization framework used in our implementation.}
	\begin{center}
		\begin{tabular}{ |c|c|c|c| }
			\hline
			\textbf{Objective} & Privacy & Fairness & Utility \\\hline
			\textbf{Quantified as} & $\epsilon$ & SPD & Accuracy \\\hline
			\makecell{\textbf{Original} \\ \textbf{output domain}} & $[0, \infty]$ & $[0, 1]$ & $[0, 1]$ \\\hline
			\textbf{Transformation} & $-\log(x)$ & \makecell{$\log(1-x) -$ \\ $\log(x)$} & \makecell{$\log(x) -$ \\ $ \log(1 - x)$} \\\hline
			\makecell{\textbf{Transformed} \\ \textbf{Output Domain}} & $[-\infty, \infty]$ & $[-\infty, \infty]$ & $[-\infty, \infty]$ \\\hline
		\end{tabular}
	\end{center}
	\label{table:transformed-objectives}
\end{table}

\clearpage

\section{Rendering of the discovered Pareto front}\label{appendix:3d-pareto-vis}

This section provides further discussion on the rendering of the Pareto fronts discovered by PFairDP. These fronts are 3-dimensional, space and hard to query for arbitrary points. This makes their visualizations a challenging task. Even a simpler problem of visualizing a complete 3D Pareto frontier is an area of active modern research with suggested methods including use of radial coordinate system \cite{Ibrahim20163DRadVisVO} and virtual reality technology \cite{Madetoja2008VisualizingMP}.

While we consider detailed research of such visualization methods to be out of scope of this work, in this section we briefly discuss a few simple plotting options, all of which can be achieved with PFairDP.

\begin{figure}[h]
	\newcommand{\scale}{0.65}
	\includegraphics[width=\scale\columnwidth]{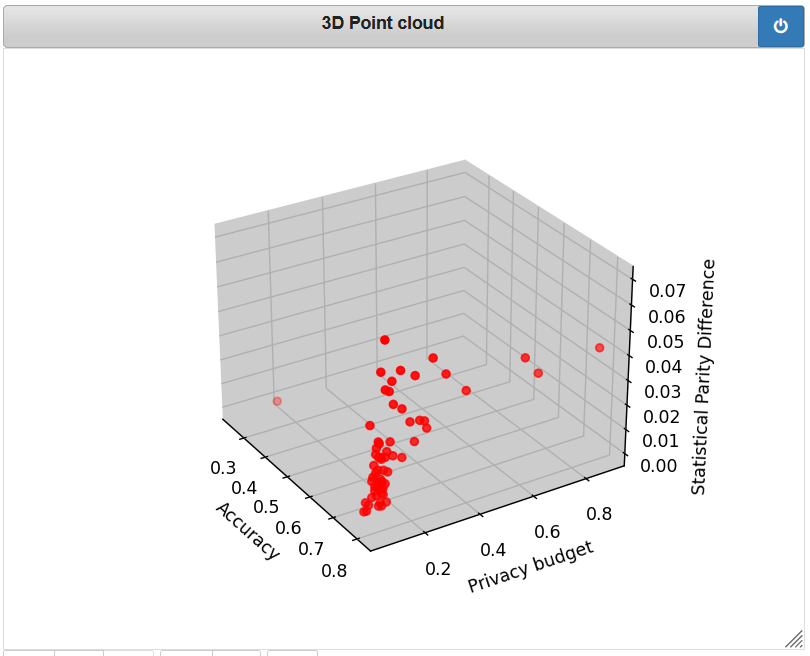}
	\caption{Screenshot of an interactive 3D point cloud produce by PFairDP.}
	\label{figure:3d_point_cloud}
\end{figure}
\begin{figure}[h]
	\newcommand{\scale}{0.65}
	\includegraphics[width=\scale\columnwidth]{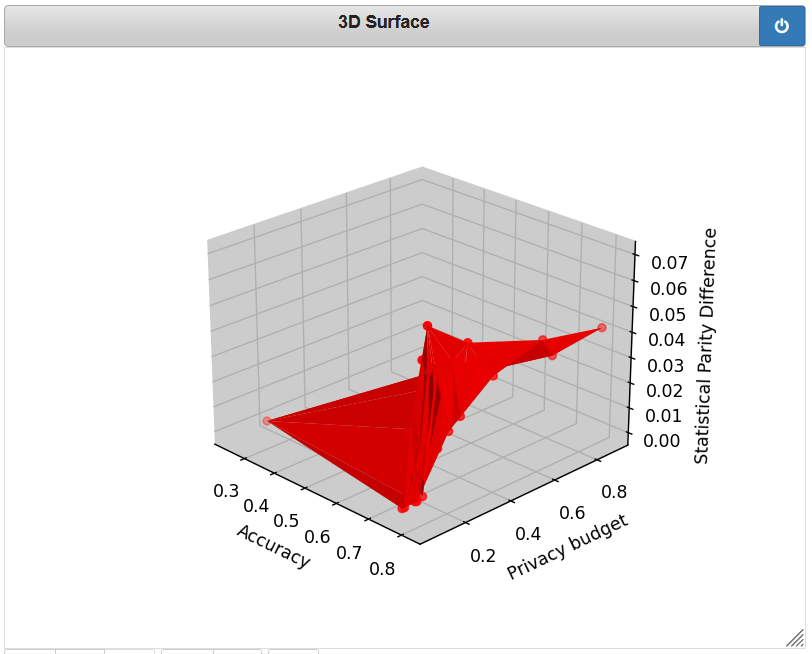}
	\caption{Screenshot of an interactive 3D surface produced by PFairDP.}
	\label{figure:3d_surface}
\end{figure}

Option 1 is the scatter plot of Pareto-optimal and optionally dominated points. This plot is very challenging to interpret statically. However with modern plotting tools like Matplotlib and Plotly users can make these plots interactive, with the ability to inspect any particular region of the entire front closer. An example of static scatter plot like that can be see on figure \ref{figure:3d_point_cloud}.

Option 2 is the 3D surface of the plot, which is a approximated triangulation on the grid of discovered points. Similarly to the option 1, this plot is hard to interpret statically, but is more practical in its interactive form. An example of this plot is given on figure \ref{figure:3d_surface}.

\begin{figure}[h]
	\newcommand{\scale}{0.45}
	\begin{subfigure}[b]{\scale\columnwidth}
		\includegraphics[width=\columnwidth]{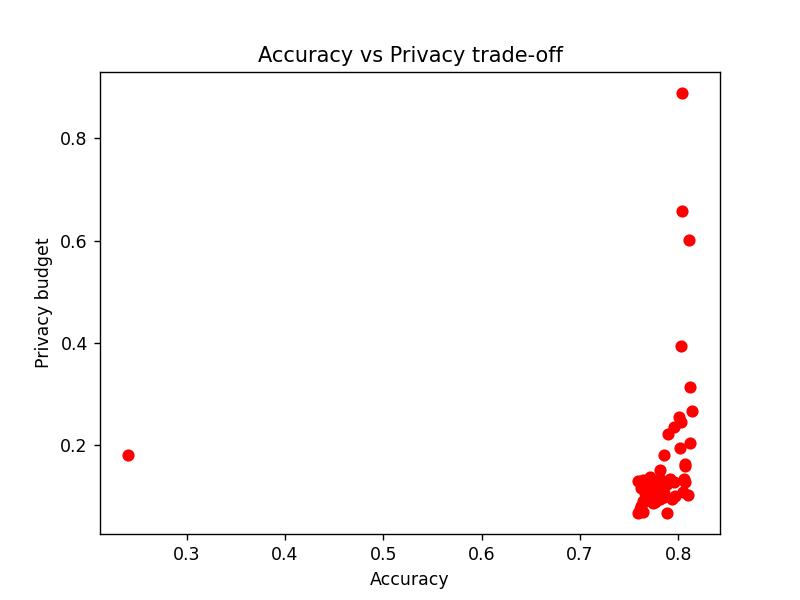}
	\end{subfigure}
	\begin{subfigure}[b]{\scale\columnwidth}
		\includegraphics[width=\columnwidth]{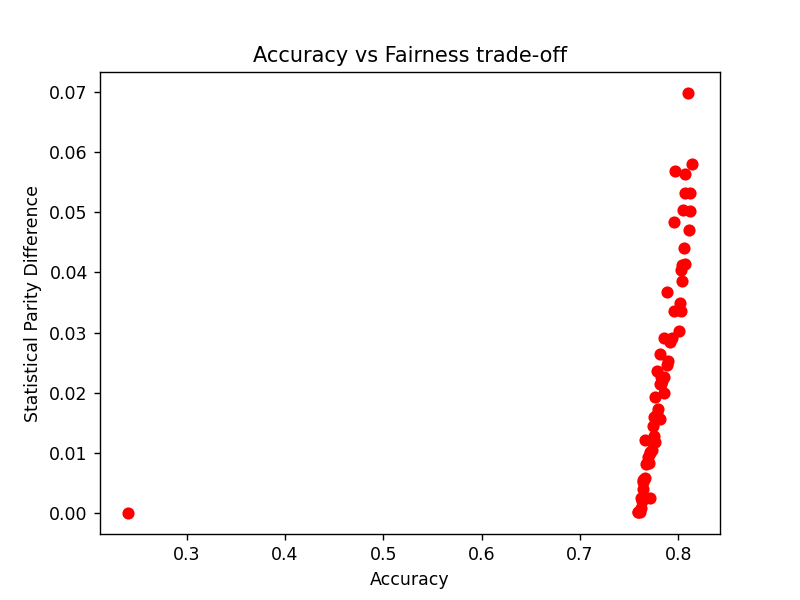}
	\end{subfigure}
	\begin{subfigure}[b]{\scale\columnwidth}
		\includegraphics[width=\columnwidth]{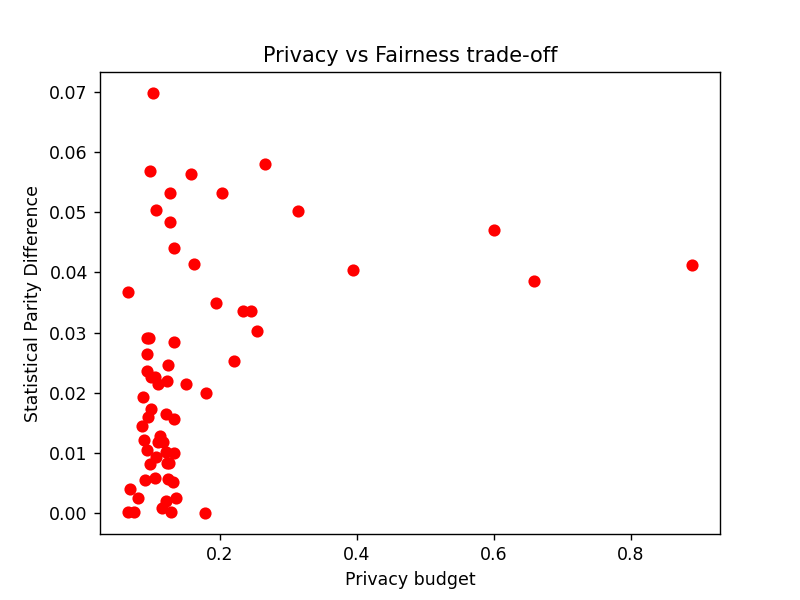}
	\end{subfigure}
	\caption{Pairwise trade-offs from the same 3D Pareto front.}
	\label{figure:pairwise-interactions}
\end{figure}

If pairwise interaction of the objectives is of interest, practitioners might find 2D cross-sections of the 3D front. While using this technique it is important to bear in mind that Pareto-optimal 3D points might not be optimal when projected to a 2D surface in the objective space. Figure \ref{figure:pairwise-interactions} gives examples of such cross-sections.
	
\end{document}